\documentclass[runningheads]{llncs}
\usepackage[T1]{fontenc}
\usepackage[misc]{ifsym}
\usepackage{graphicx}
\newcommand{\indep}{\perp \!\!\! \perp}
\usepackage{bbm}

\usepackage{cite}
\usepackage{amsmath,amssymb,amsfonts}
\usepackage{graphicx}
\usepackage{textcomp}
\usepackage{xcolor}
\usepackage[square,numbers]{natbib}
\usepackage{colortbl}
\usepackage{array}
\newcolumntype{P}[1]{>{\centering\arraybackslash}p{#1}}
\definecolor{lightgray}{RGB}{212,212,212}
\usepackage{multirow}
\usepackage{tabu}                      
\usepackage{booktabs}                  
\usepackage{pifont}
\usepackage{algorithm}
\usepackage{algpseudocode}
\usepackage[inline]{enumitem}
\usepackage{float}
\usepackage{placeins}

\usepackage[font=small,skip=0pt]{caption}
\newcommand{\cmark}{\ding{51}}%

\usepackage{tikz} 
\usepackage[utf8]{inputenc}

\usepackage{hyperref}
\usepackage{cleveref}
\newcommand{\ind}{\perp\!\!\!\!\perp}

\def\BibTeX{{\rm B\kern-.05em{\sc i\kern-.025em b}\kern-.08em
    T\kern-.1667em\lower.7ex\hbox{E}\kern-.125emX}}
\begin{document}
\title{Uncovering Causal Relation Shifts in Event Sequences under Out-of-Domain
Interventions}
\titlerunning{Causal Relation Shifts in Event Sequence}
%

\author{Kazi Tasnim Zinat\Letter\inst{1}\orcidID{0000-0001-7914-5955} \and
Yun Zhou\inst{2} \and
Xiang Lyu\inst{2}\orcidID{0009-0000-1398-3374} \and
Yawei Wang\inst{2} \and
Zhicheng Liu\inst{1}\orcidID{0000-0002-1015-2759} \and
Panpan Xu\inst{2}}
\authorrunning{Zinat et al.}
%
\institute{University of Maryland, College Park, MD 20740, USA \email{\{kzintas, leozcliu\}@umd.edu}
\and
Amazon Web Services
\email{\{yunzzhou,xianglyu,yawenwan,xupanpan\}@amazon.com}}
\maketitle              
\begin{abstract}
Inferring causal relationships between event pairs in a temporal sequence is applicable in many domains such as healthcare, manufacturing, and transportation. Most existing work on causal inference primarily focuses on event types within the designated domain, without considering the impact of exogenous out-of-domain interventions.  In real-world settings, these out-of-domain interventions can significantly alter causal dynamics. 
To address this gap, we propose a new causal framework to define average treatment effect (ATE), beyond independent and identically distributed (i.i.d.) data in classic Rubin's causal framework, to capture the causal relation shift between events of temporal process under out-of-domain intervention.
We design an unbiased ATE estimator, and devise a Transformer-based neural network model to handle both long-range temporal dependencies and local patterns while integrating out-of-domain intervention information into process modeling.
Extensive experiments on both simulated and real-world datasets demonstrate that our method outperforms baselines in ATE estimation and goodness-of-fit under out-of-domain-augmented point processes. The supplementary materials are available \href{https://osf.io/trx3y/?view_only=bd89422cf70a41b19906a775261b07f0}{here}.

\end{abstract}

%
%
%
\section{Introduction}
Multivariate event sequences are prevalent across diverse domains, capturing time-stamped data from shared environments or subjects, such as Electronic Health Records \citep{xiao2018opportunities} and industrial maintenance records \citep{Zhou2023}. Real-world data generation is often implicitly conditioned on unobserved out-of-domain variables, leading to generalization failures due to distribution shifts \citep{koh2021wilds, zheng2020out}.

Causal relation analysis within event sequences aims to quantify how changes in one event influence another \citep{gao2021causal, noorbakhsh2022counterfactual}. The Rubin Causal Model \citep{imbens2015causal} characterizes causal relations through the average treatment effect (ATE), representing the average difference between treatment \& control potential outcomes. However, existing work focuses on causal relations within designated domains, overlooking the influence of known exogenous out-of-domain interventions that can induce causal relationship shifts. For instance, while meal ingestion increases blood glucose, insulin injection (an out-of-domain intervention) can counteract this effect \citep{campbell2014metabolic}.\looseness-1

To address this gap, we propose a novel approach to detect causal relation shifts in temporal processes under out-of-domain interventions. Our framework extends ATE to account for unique intervention scenarios, enabling comparison across such contexts. To our knowledge, this is the first work to explicitly model out-of-domain intervention impact on causal relations in temporal event data.\looseness-1

Our main contributions are multi-fold. First, we develop a theoretical framework of ATE that moves beyond i.i.d. assumptions to explicitly condition on intervention states and temporal dependencies. Second, we develop a propensity score-based treatment effect estimator to mitigate confounder bias. We justify its estimation consistency under our new ATE framework. Third, we propose a new Transformer architecture that captures intervention-induced temporal pattern changes. The architecture consists of several new network components: 1) a novel out-of-domain intervention embedding mechanism that enables direct influence modeling on event sequences; 
b) a weighted combination module that adaptively balances intervention and event embeddings;
3) a hybrid Transformer-CNN structure to simultaneously capture global dependencies and local temporal patterns specific to intervention effects, 4) a multi-objective loss function that jointly optimizes for intensity estimation and event type prediction. Fourth, we provide comprehensive experimental validation on simulated and real-world datasets, demonstrating improved performance in both causal effect estimation and temporal process representation compared to existing methods.

\section{Related Work}
Point processes \cite{daley2003introduction} model labeled event sequences, with Hawkes processes \cite{hawkes1971point} capturing excitation/inhibition effects of past events on current occurrences. Neural approaches include RNNs \cite{du2016recurrent, xiao2017modeling} for handling time and labels simultaneously, and Transformers \cite{zuo2020transformer, zhang2022temporal} for long-range dependencies \cite{shchur2021neural}.

Causal analysis compares potential treatment and control outcomes using the Rubin Causal Model \citep{schulam2017reliable}, with ATE as the primary metric. Propensity scores mitigate confounding bias in observational studies \citep{imbens2015causal}, while deep learning advances causal estimation \citep{shi2019adapting, melnychuk2022causal}. Temporal causal inference extends these concepts through Granger causality \citep{zhang2020cause} and point process-based causal modeling \citep{gao2021causal, noorbakhsh2022counterfactual}. A detailed discussion with other methods is provided in 
\cref{appendix:SOTA}
 \section{Out-of-Domain Intervention Augmented Causal Inference}
\subsection{Notation}
Let us consider a set of $n$ individual sequences \( \{\mathbf{s}_1, , \ldots, \mathbf{s}_n \} \). 
In a generic event sequence scenario, for a single sequence $\mathbf{s}_k$, the observation we collect for the ${i}$-th event can be expressed as a tuple ${(e_{k,i}, t_{k,i})} $. Here, $ {t_{k,i}} $ represents the timestamp of the $ {i} $-th event in sequence $ \mathbf{s}_k $, and $ {e}_{k,i}$ denotes the corresponding event type which belongs to a set of events $\mathbbm{E}$. 
The count of observed events of sequence \( \mathbf{s}_k \) is denoted by \( {L_{k}} \).
The time duration of each sequence is $T$.
We  categorize the events based on their types: cause events $c$,  outcome events $o$, out-of-domain interventions $v$, and other measured events $\mathbf{x}$. 
\subsection{Temporal Point Process}


We focus on temporal point processes, where both event type and timestamp are observed. Hawkes processes, known for their self-excitation property, are commonly used to model real-world event sequences \citep{mei2017neural, zhang2022temporal}. Specifically, for an \(n\)-dimensional Hawkes process, the conditional intensity function (CIF) of outcome event \(o_i\) at time \(t\) is expressed as:

\begin{equation} 
\setlength{\abovedisplayskip}{-5pt}
\setlength{\belowdisplayskip}{-5pt}
\label{eq:CIF}
\textstyle
       {\lambda}{(t)} = {\mu} (t)  + \sum_{k=1}^{n} \sum_{  t_{k,i}<t} {\phi}(t-t_{k,i}),  
\end{equation}
where \({\mu}\) denotes the baseline intensity function, 
\({\phi}\) is the excitation function capturing the influence of past events on  outcome event.
Note that $\lambda (t)$ is not observable. In this work, we estimate it from sequence data.



\subsection{Out-of-Domain Intervention Augmented Causal Framework}\label{subsec:causaldisc}

We extend the Rubin causal framework to incorporate out-of-domain interventions, leveraging two key concepts.

First, we adopt the notion of process independence from graphical models \citep{Didelez_2005}, to establish direct causal relationships in multivariate point processes. For event variables $(x, y, z)$, $x$ is process independent of $y$ given $z$ if CIF of $x$ is not functionally dependent on the history of $y$ given the history of $z$. Consequently, a set of events $\mathbb{X}$ is considered a direct cause of event $y$ if $y$ is process independent of all other events given $\mathbb{X}$. We exclude the trivial case where  $y$ is constant over $\mathbb{X}$.\looseness -1

Second, we introduce the concept of proximal history, a time-based simplification widely used in point process literature \citep{Bhattacharjya_2018}. Proximal history posits that cause events only within a recent time window  influences the outcome event, allowing earlier history to be disregarded.

Combining these notions, we establish a causal framework and define ATE under out-of-domain interventions, drawing inspiration from \citep{gao2021causal} while adapting the theoretical assumptions for unbiasedness to accommodate out-of-domain interventions. We further incorporate these interventions into the propensity score-adjusted estimator, modifying the score definition. 

\begin{definition}
For an event tuple $(c, o , v)$, binary cause variable $c_t^w$ at time $t$ indicates whether cause event $c$ has occurred at least once in the time window $[t-w,t)$. Similarly, binary out-of-domain intervention variable $v_t^w$ indicates whether out-of-domain intervention $v$ occurred within the same window.
The potential outcome variable $\lambda^{(c_t^w,v_t^w)} (t)$ denotes the CIF of outcome event $o$ at time $t$, given the values of $c_t^w$ and $v_t^w$. 
The binary vector $\mathbf{x}_t^w$ captures the occurrence of all other observed events
in the time window $[t-w,t)$. 
The ATE of cause $c_t^w$ on outcome $\lambda(t)$ under out-of-domain intervention $v_t^w$ is defined as:
\begin{equation*}
\setlength{\abovedisplayskip}{2pt}
\setlength{\belowdisplayskip}{2pt}
\textstyle
  \tau (v_t^w) = \mathbb{E} \bigg [ \frac{1}{T} \int_{0}^T  \{\lambda^{(1,v_t^w)} (t) - \lambda^{(0,v_t^w)} (t)\} dt \bigg ].  
\end{equation*}
Unlike standard causal settings,
$\lambda(t)$ is latent and estimated from event type and timestamp observations.
\end{definition}


Cause and out-of-domain interventions may exhibit multiple occurrences prior to outcome within the time window. To focus on detecting causal relation shifts under interventions, we simplify the framework to study the treatment effect of occurrence, rather than count, as an initial step. Our real-world datasets validate this binary simplification. We defer continuous and ordinal cause and intervention modeling to future work. While not explicitly defined, process independence is crucial for linking the new ATE definition to causal relation shifts.


\begin{theorem}
\label{thm1}
If event $c$ is the direct cause of event $o$ when out-of-domain intervention variable $v_t^w=1$, and not if $v_t^w=0$, then $\tau (1)\neq 0$ and $\tau (0) =0.$
\end{theorem}

\Cref{thm1} demonstrates that our new ATE effectively characterizes causal relations under out-of-domain interventions; the proof is provided in \cref{appendix:proofs_thm1}.

\subsection{Treatment Effect Estimation}

In point process causal studies, cause event occurrences are typically not randomized, leading to potential imbalances between treatment groups. To address this, we propose a propensity score, adapted for our out-of-domain intervention augmented ATE framework, building on the established use of propensity scores for confounder adjustment in classical observational studies.


\begin{definition}
The propensity score at time $t$ for out-of-domain intervention variable $v_t^w$  is 
\begin{equation*}
\setlength{\abovedisplayskip}{-5pt}
\setlength{\belowdisplayskip}{-5pt}
\textstyle
  e_t^w(v) = \mathbb{P} \{ c_t^w = 1 , v_t^w = v | \mathbf{x}_t^w\}, v \in \{0,1\}.  
\end{equation*}
\end{definition}


Our novel propensity score, $e_t^w(v)$, incorporates both cause and out-of-domain intervention variables, conditioned on the proximal history of all other observed events. By treating binary out-of-domain interventions as alternative treatment assignments, rather than effect heterogeneity factors, we simplify the theoretical framework for estimation consistency.

We demonstrate that, with minor modifications to standard observational study assumptions to accommodate point process data and out-of-domain interventions, $e_t^w(v)$ effectively mitigates confounding bias, yielding an unbiased effect estimator.

\noindent \textbf{Assumption 1 (SUTVA):}
For each assignment pair $(c_t^w, v_t^w)$ and any $t$, there is only a single version of population outcome $\lambda^{(c_t^w, v_t^w)} (t)$, and the time window receives the assignment will not affect the outcome of other time windows.\looseness -1  

The assumption is fundamental in causal inference. 
It ensures we can leverage observations under treatment assignment to infer potential outcomes.

\noindent\textbf{Assumption 2 (unconfoundedness):}
For any $t$, we have
\begin{equation*}
\textstyle
\setlength{\abovedisplayskip}{3pt}
\setlength{\belowdisplayskip}{3pt}
\big \{\lambda^{(0,0)} (t), \lambda^{(0,1)} (t), \lambda^{(1,0)} (t), \lambda^{(1,1)} (t) \big \}\indep  \big(c_t^w, v_t^w\big) | \mathbf{x}_t^w.      
\end{equation*}
This assumption guarantees measured covariates sufficiently balance treatment groups and adjust for confounding bias across different out-of-domain interventions. While untestable, it is commonly postulated in observational studies.\looseness-1

\noindent \textbf{Assumption 3 (overlap):}
There exists a constant $\epsilon$ such that 
$\epsilon <  e_t^w(v) < 1-\epsilon,\forall t, \forall v \in \{0,1\}.$ 

This assumption ensures sufficient observations across treatment groups under varying out-of-domain interventions for accurate estimation, and is verifiable in practice. Our real-world datasets satisfy this condition.


To simplify theoretical justification, we treat out-of-domain interventions as a second binary treatment variable, consistent with our framework's focus on the occurrence of cause and intervention events. This approach avoids the need for complex assumptions regarding potential outcomes and the impact of out-of-domain interventions on cause events, which would be overly intricate for our current framework and real-data applications.

We leverage propensity score $e_t^w(v)$ to construct weight 
\begin{equation*}
\setlength{\abovedisplayskip}{3pt}
\setlength{\belowdisplayskip}{3pt}
\textstyle
  \alpha_t^w (v) = \frac{ \mathbbm{1} \{c_t^w=1, v_t^w=v\}}{e_t^w(v)} - \frac{\mathbbm{1} \{c_t^w=0, v_t^w=v\}}{1-e_t^w(v)}.  
\end{equation*}

We weight the outcome $\lambda(t)$ to derive the inverse probability weighting (IPW) estimator of ATE,  
\begin{equation}
\setlength{\abovedisplayskip}{0pt}
\setlength{\belowdisplayskip}{2pt}
\textstyle
\label{eq:IPTW}
   \hat{\tau}(v) = \mathbb{E} \bigg [ \frac{1}{T} \int_{0}^T \alpha_t^w (v) \lambda(t)  dt \bigg ]. 
\end{equation}

We show that the IPW estimator is unbiased. 

\begin{theorem}    
Under Assumption 1-3, we have $\mathbb{E} [ \hat{\tau}(v)] = \tau(v), \forall v \in \{0,1\}.$ 
The propensity score can be estimated by event duration ratio,  
\begin{equation} 
\label{eq:propensity_score}
\setlength{\abovedisplayskip}{3pt}
\setlength{\belowdisplayskip}{4pt}
\textstyle
\hat{e}_t^w(v) = \frac{\sum_{k=1}^n \sum_{i=1}^{L_{k}} \int_{t_{i-1}}^{t_i} \mathbbm{1} \{ c_t^w = 1 , v_t^w = v ,\mathbf{x}_t^w \} dt }{
\sum_{k=1}^n \sum_{i=1}^{L_{k}} \int_{t_{i-1}}^{t_i} \mathbbm{1} \{ \mathbf{x}_t^w \} dt
}.
\end{equation}
\end{theorem}

We obtain the effect estimate by plugging $\hat{e}_t^w(v)$ into the weight $\alpha_t^w(v)$  in \cref{eq:IPTW}. The outcome $\lambda(t)$ is estimated using a Transformer-based model on sequence data; see \cref{appendix:proofs_thm2}
for the proof.

\section{Transformer-based Process Model}



We propose a neural network model for learning temporal point processes and estimating the conditional intensity function (CIF) $\lambda(t)$  of outcome events under both cause and out-of-domain interventions. A key innovation is the integration of out-of-domain interventions into the input representation.

\begin{figure}[h]
\centering
\includegraphics[width=0.45\textwidth]{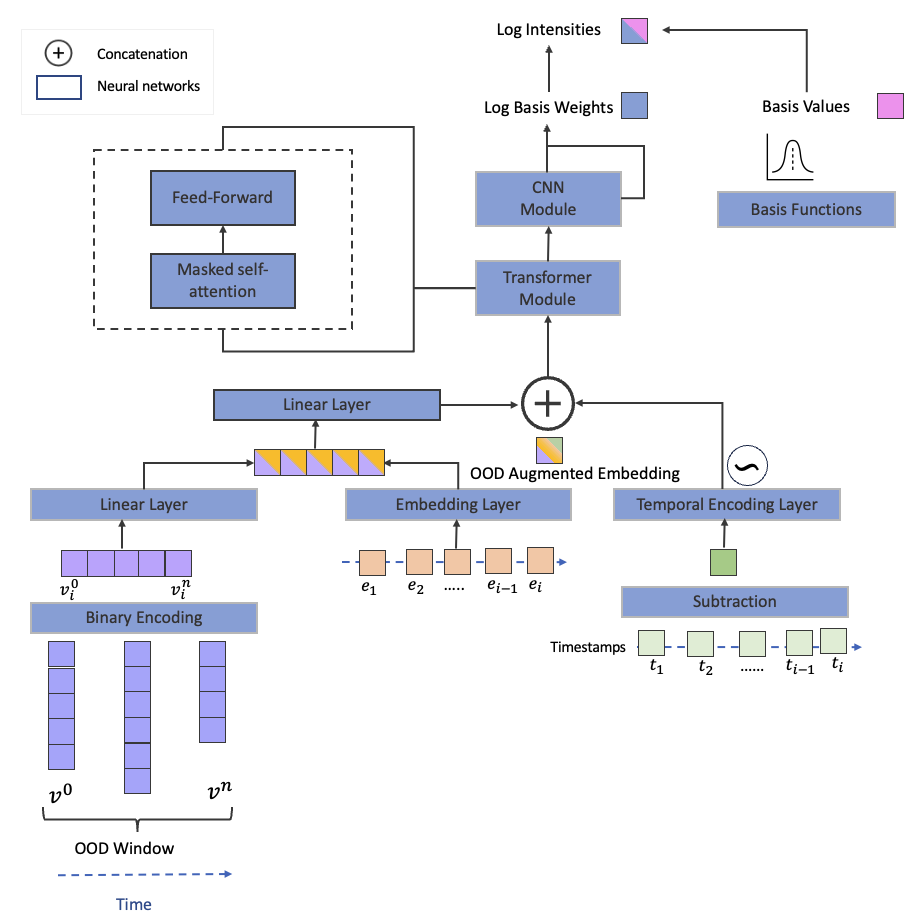}    
\caption{
Our neural network architecture for learning temporal point processes with out-of-domain interventions. It combines out-of-domain intervention-augmented ($V_i$) event embeddings ($e_i$) with positional encoding ($\Delta t$), uses Transformer and CNN modules for pattern extraction,  and estimates the CIF $\lambda(t)$ for event prediction
}
\label{fig:NNarchitecture}
\setlength\abovecaptionskip{-70mm}
\setlength\belowcaptionskip{-10mm}
\end{figure}

Our neural network model, illustrated in \cref{fig:NNarchitecture}, adopts a hybrid Transformer-CNN architecture to capture long-range temporal dependencies and local patterns shaped by intervention dynamics. 
A self-attention block \citep{vaswani2017attention} models event interactions via attention weights, with multi-head attention enhancing expressiveness by capturing diverse relational patterns. 
To incorporate temporal context, we embed relative time differences $t_i - t_{i-1}$ using temporal positional encodings \citep{zuo2020transformer}. 
This architecture is used to model the logarithm of the outcome CIF, $\lambda(t)$, which is central to ATE estimation.


Our model input comprises three encoded components: 1) binary indicators of out-of-domain interventions $v$ occurring within the time window; 2) dense embeddings of event types (cause, outcome, and other observed events); and 3) relative event times, $t_i - t_{i-1}$, encoded using trigonometric positional encodings to capture temporal positions. We combine intervention and event type embeddings via a weighted sum to form a unified, intervention-aware event representation, which is then augmented with temporal positional encodings. The model outputs the conditional intensity function (CIF) for each event and timestamp, expressed as a sum over basis functions.

We optimize our model using three criteria combined into a single loss function.
raining is performed on batches of sequences $B = \{s_1, s_2, ..., s_b\}$ where $b$ is the batch size. For each batch, we compute the average loss across all sequences.

The first criterion is negative log likelihood (NLL) of event observations within the sequences. 
\begin{equation}
\setlength{\abovedisplayskip}{0pt}
\setlength{\belowdisplayskip}{2pt}
\label{eq:nll}
\textstyle
    L_{NLL} = \frac{1}{b}\sum_{k=1}^{b} \left[-\sum_{i=1}^{L_k} \log \lambda_{e_{k,i}}(t_{k,i}) + \sum_{e \in \text{outcome}} \int_0^T \lambda_e(t)dt\right],
\end{equation}
where $\lambda_{e_{k,i}}(t_{k,i})$ is the conditional intensity function for event $e_{k,i}$  at timestamp $t_{k,i}$, and $L_k$ is the length of sequence $k$. The first term represents the log-likelihood of observed events, while the second accounts for the probability of no events occurring in the remaining time intervals.


The second criterion is a prediction loss based on cross-entropy that measures how well the model predicts event types.
\begin{equation}
\setlength{\abovedisplayskip}{1.5pt}
\setlength{\belowdisplayskip}{1pt}
\label{eq:celoss}    
\textstyle
\mathcal{L}_{CE} = - \frac{1}{b} \sum_{k=1}^{b}\sum_{i=1}^{L_k}\sum_{e \in \mathbb{E}} y_{k,i,e} \log(p_{k,i,e}),
\end{equation}
where $\mathbb{E}$ is the set of event classes,
 $y_{k,i,e}$ is the true binary indicator (0 or 1) of whether event $e$ is the correct classification for $i$-th event in $\mathbf{s}_k$, $p_{k,i,e}$ is the predicted probability that $i$-th event belongs to class $e$.


The third criterion is L2 regularization term that penalizes large basis weights to prevent overfitting.
\begin{equation}
\setlength{\abovedisplayskip}{2pt}
\setlength{\belowdisplayskip}{2pt}
\label{eq:L2}
\textstyle
L_{reg} =   \frac{1}{b} \sum_{k=1}^{b}  \sum_{i=1}^{L_k} \sum_{l=1}^\mathcal{B} \exp(w_{l,k,i})^2,
\end{equation}
where $w_{l,k,i}$ represents the log basis weights at time $t_{k,i}$ for basis function $l$, $\mathcal{B}$ denotes the total number of basis functions.
The regularization prevents the basis weights from growing too large, which could lead to numerical instabilities.

Our final batch loss function combines the three criteria, \eqref{eq:nll}, \eqref{eq:celoss} and \eqref{eq:L2}, 
\begin{equation}
\setlength{\abovedisplayskip}{2pt}
\setlength{\belowdisplayskip}{2pt}
\textstyle
L_{batch} =  L_{NLL}  + \alpha L_{CE} + \beta L_{reg}.  
\end{equation}

The hyperparameters $\alpha$ and $\beta$ control the relative importance of the cross-entropy loss and regularization term, respectively.
We minimize this batch loss using the Adam optimizer and evaluate convergence on a held-out validation set. Model selection is performed via 5-fold cross-validation. Additional hyperparameter and implementation details are provided in \cref{appendix:implementation details}.

\section{Numerical Study}

We evaluated our method on both simulated and real-world data, comparing it against the CAUSE model \citep{zhang2020cause} for ATE estimation and out-of-domain intervention-augmented process learning. CAUSE, which combines point processes and attribution methods for Granger causality inference, estimates the CIF without accounting for out-of-domain interventions. A comparative discussion with other models is provided in the \cref{appendix:SOTA}. 

\subsection{Simulated Data}

We generated synthetic event sequences with injected out-of-domain interventions, modeled as random events influencing specific cause–outcome pairs. Each intervention was assigned a random occurrence probability and time window,
with dynamic injection based on these probabilities. Additional simulation details are provided in the \cref{appendix:simulation}.

We simulated three intervention impact types, increasing in complexity and coverage of out-of-domain intervention impact: 
\begin{enumerate}[leftmargin=*,  noitemsep, topsep=1pt]
    \item \textbf{No out-of-domain intervention (No OOD)}: we do not impose out-of-domain intervention on process, which also can serve as sanity check. 
    \item \textbf{Baseline out-of-domain intervention (Baseline}): we modify the baseline intensity $\mu (t)$ of an outcome event CIF $\lambda (t)$ under out-of-domain intervention shown in Eq. \ref{eq:CIF}. We implement $30$ out-of-domain interventions per sequence. 
    \item \textbf{All out-of-domain intervention (All impact)}: we alter baseline intensity $\mu (t)$, the influence of cause and other observed events through $\phi(\cdot)$ on outcome event CIF $\lambda (t)$ under out-of-domain intervention. We implement $30$ out-of-domain interventions per sequence.
\end{enumerate}

We generate $1,000$ sequences for each intervention impact type, with sequence lengths drawn from a Poisson distribution (mean = $500$). Each sequence includes $30$ distinct event types: $20$ used as causes and covariates, and the remaining $10$ are designated as outcomes.

We compare our method with CAUSE in ATE estimation and process fitting in simulated data. 
The performance measures we adopt for ATE estimation are 
\begin{enumerate}[itemsep=0pt, topsep=1pt, label={}, leftmargin=*]
\item \textbf{Bias}: the average of absolute difference between ATE estimate and true ATE across repetitions. 
\item \textbf{Variance}: the variance of ATE estimate across repetitions. 
\item \textbf{Mean squared error (MSE)}: the average of squared difference between ATE estimate and true ATE across repetitions. 
\end{enumerate}

Lower values indicate better ATE estimation. As shown in \cref{tbl:simulated_ATEestimation}, our method significantly reduces bias across all scenarios, with slight increase in variance. This trade-off is beneficial, as it results in lower MSE in most cases.

\begin{table}[h]
\fontfamily{ptm}\selectfont\scriptsize
\caption{Simulated data ATE estimation performance comparison between our method and CAUSE for Baseline and All impact interventions. Lower values are better.
} \label{tbl:simulated_ATEestimation}
\centering
{\begin{tabu}{p{2cm}p{1.5cm}p{1.3 cm}p{1.3 cm}p{1.3cm}p{1.3cm}p{1.3cm}p{1.3cm}}
\toprule		
\multirow{2}{*}{\parbox{2cm}{{\centering\arraybackslash} Out-of-Domain\\Intervention}} &
\multirow{2}{*}{\parbox{2.5cm}{Intervention\\Status}} &
\multicolumn{2}{c}{Bias} &
\multicolumn{2}{c}{Variance} &
\multicolumn{2}{c}{MSE}\\
\cline{3-8}
\multicolumn{1}{c}{} &
\multicolumn{1}{c}{} &
\multicolumn{1}{c}{Ours} & 
\multicolumn{1}{c}{Cause} &
\multicolumn{1}{c}{Ours} & 
\multicolumn{1}{c}{Cause} &
\multicolumn{1}{c}{Ours} & 
\multicolumn{1}{c}{Cause}\\
\midrule
\multirow{2}{*}{Baseline}  
 & 0 & \textbf{0.0155} & 0.0522 &  0.0141  &  \textbf{0.00001} & \textbf{0.003} & 0.0046 \\
\cline{2-8}
 & 1 & \textbf{0.1561} & 0.262 &  \textbf{0.0002} & \textbf{0.0002}  & \textbf{0.027} & 0.0708\\
\hline
\multirow{2}{*}{All Impact}   
 & 0 & \textbf{0.0209} & 0.0494 & 0.0027  & \textbf{0.00001}  & \textbf{0.0038} & 0.0042 \\
\cline{2-8}
 & 1 & \textbf{0.1615} & 0.2192 & 0.0008  &  \textbf{0.0002} & \textbf{0.0283} & 0.0498 \\
\bottomrule
\end{tabu}}
\end{table}

We use the following performance measures for process fitting, 
\begin{enumerate}[leftmargin=*, label={}, itemsep=0pt, topsep=0pt]
\item \textbf{Negative log likelihood (NLL)}: the NLL of fitted process model.
\item \textbf{Root mean squared error (RMSE)}: the square root of the average of squared outcome event occurrence time prediction error. 
\item \textbf{Mean absolute error (MAE)}: the average of absolute outcome event prediction error. 
\end{enumerate}

Averaged across repetitions, \cref{tbl:simulated_processfit} shows that our model consistently outperforms CAUSE across all scenarios by achieving lower NLL, RMSE, and MAE; demonstrating superior modeling of event distributions and predictive accuracy, even under varying intervention complexities.

\begin{table}[t]
\fontfamily{ptm}\selectfont\small
\caption{
Simulated data process fitting performance comparison (NLL, RMSE, MAE) between our method and CAUSE with varying Out-of-Domain interventions. Lower values are better.
}
\label{tbl:simulated_processfit}
\centering
        {\scriptsize \begin{tabu}{ p{2.5cm}  p{1.5 cm}p{1.5 cm}p{1.3cm}p{1.3cm}p{1.3cm}p{1.3cm} }
\toprule		

\multirow{2}{*}{\parbox{2cm}{{\centering\arraybackslash} Out-of-Domain\\Intervention}} &
\multicolumn{2}{c}{NLL} &
\multicolumn{2}{c}{RMSE} &
\multicolumn{2}{c}{MAE}\\
\cline{2-7}

\multicolumn{1}{c}{} &
\multicolumn{1}{c}{Ours} & 
\multicolumn{1}{c}{Cause} &
\multicolumn{1}{c}{Ours} & 
\multicolumn{1}{c}{Cause} &
\multicolumn{1}{c}{Ours} & 
\multicolumn{1}{c}{Cause}\\
\midrule

No OOD & \textbf{1009.06} & 2456.53 & \textbf{1.78}  &  3.89 & \textbf{0.89} & 2.60 \\
\hline
Baseline  & \textbf{852.48} & 2438.36 & \textbf{2.1}  & 3.84  & \textbf{1.04} & 2.56 \\
\hline
All Impact  & \textbf{1299.61} & 2443.81 &  \textbf{2.18} &  3.86 & \textbf{1.17} & 2.58 \\
\bottomrule
\end{tabu}}
\end{table}

\subsection{Real-World Data}

We evaluated our method on two real-world datasets:

\noindent \textbf{Predictive Maintenance}\citep{Azure}: This dataset contains hourly sensor readings (voltage, rotation, pressure, vibration) from 100 machines. Proactive maintenance events (scheduled component replacements) were modeled as out-of-domain interventions, while reactive maintenance events (failures requiring unscheduled replacements) served as outcomes. Continuous sensor readings were discretized into 625 bins based on mean deviation, with each bin treated as a potential cause event and others as time-varying covariates.

\begin{table}[b]
\fontfamily{ptm}\selectfont
\scriptsize
\caption{Predictive maintenance data process fitting performance (NLL, RMSE, MAE) comparison between our method and CAUSE, with proactive maintenance as out-of-domain intervention. Lower values are better.
} \label{tbl:maintenance_processfit}
\centering
        \newcolumntype{C}[1]{>{\centering\arraybackslash}p{#1}}
        {\small \begin{tabu}{ C{0.1\linewidth}  C{0.1\linewidth}  C{0.1\linewidth}  C{0.1\linewidth}  C{0.1\linewidth}  C{0.1\linewidth} }

\multicolumn{2}{C{0.225\linewidth}}{NLL} &
\multicolumn{2}{C{0.225\linewidth}}{RMSE} &
\multicolumn{2}{C{0.225\linewidth}}{MAE}\\
\midrule 
\multicolumn{1}{P{0.09\linewidth}}{Ours} & 
\multicolumn{1}{P{0.09\linewidth}}{Cause} &
\multicolumn{1}{P{0.09\linewidth}}{Ours} & 
\multicolumn{1}{P{0.09\linewidth}}{Cause} &
\multicolumn{1}{P{0.09\linewidth}}{Ours} & 
\multicolumn{1}{P{0.09\linewidth}}{Cause}\\
\midrule 
  300.66 & \textbf{288.84} & \textbf{419.38}  & 3501.14  & \textbf{59.88} & 628.37 \\
\bottomrule
\end{tabu}}
\end{table}

\noindent \textbf{Diabetes} \citep{misc_diabetes_34}: This dataset includes lab events from 70 patients, such as insulin injections, glucose levels, meal consumption, physical activity, and hypoglycemia symptoms. We focused on glucose as the outcome and selected cause and OOD intervention events based on established medical knowledge. For example, in analyzing insulin-mediated effects of meals on glucose, we treated meal as the cause, insulin as the out-of-domain intervention, and activity and hypoglycemia symptoms as covariates. A baseline experiment without interventions was also conducted to assess model robustness.

\begin{table}[t]
\fontfamily{ptm}\selectfont\small
    \caption{ATE-estimated causal relations in diabetes dataset under Out-of-domain interventions, validated by medical literature.}
    \label{tbl:diabetes_ATEestimation}
    \centering
    {\scriptsize \begin{tabu}{P{0.16\linewidth} | P{0.12\linewidth} | P{0.15\linewidth} |  P{0.2\linewidth} |  P{0.36\linewidth}}
    \toprule
          {Cause}&   {Outcome}&  {Out-of-Domain Intervention}& {Covariate}& {Our Conclusion \& Literature Evidence} \\
         \midrule
         More than usual meal ingestion &  Blood glucose decrease &  Insulin & Activity, hypoglycemia symptopm &  \textcolor{green}{\cmark}  Insulin injection, even when more than usual meal is ingested, may help to decrease blood glucose \citep{campbell2014metabolic} \\
         \hline
         Hypoglycemia symptom &  Blood glucose decrease &  NPH insulin & Meal, activity &  \textcolor{green}{\cmark} Insulin injection, when hypoglycemic symptom is observed, may cause blood glucose to decrease further \citep{lee2014so} \\
         \hline
         Insulin &  Blood glucose decrease &  Typical or more than usual activity & Meal, hypoglycemia symptom & \textcolor{green}{\cmark} More activity increases the effect of insulin, boosting blood glucose decrease \cite{lawrence1926effect}\\
         \bottomrule
          
    \end{tabu}}
\end{table}

Since ground truth ATEs and prior benchmarks are unavailable for the predictive maintenance dataset, we evaluated process fitting performance. As shown in \cref{tbl:maintenance_processfit}, our model significantly outperformed CAUSE in future event prediction:achieving an 88\% reduction in RMSE and a 90\% reduction in MAE, while maintaining comparable distribution modeling (only 4\% higher NLL). These improvements highlight the value of explicitly modeling out-of-domain interventions in capturing complex real-world dynamics.

\begin{table}[t]
\caption{Process fitting performance comparison between our method and CAUSE on the diabetes dataset, with activity and insulin as out-of-domain interventions. Sanity check: no intervention. Lower NLL, RMSE, and MAE indicate better performance.
} \label{tbl:diabetes_processfit}
\centering
       {\scriptsize \begin{tabu}{ p{2.5cm}  p{1.5 cm}p{1.5 cm}p{1.5cm}p{1.5cm}p{1.5cm}p{1.5cm} }
\toprule		
\multirow{2}{*}{\parbox{2cm}{{\centering\arraybackslash} Out-of-Domain\\Intervention}} &
\multicolumn{2}{c}{NLL} &
\multicolumn{2}{c}{RMSE} &
\multicolumn{2}{c}{MAE}\\
\cline{2-7}
\multicolumn{1}{c}{} &
\multicolumn{1}{c}{Ours} & 
\multicolumn{1}{c}{Cause} &
\multicolumn{1}{c}{Ours} & 
\multicolumn{1}{c}{Cause} &
\multicolumn{1}{c}{Ours} & 
\multicolumn{1}{c}{Cause}\\
\midrule
Sanity Check  & \textbf{613.21} & 619.74 & \textbf{15644.33}  & 15718.8  & \textbf{2104.87} & 2239.53 \\
\hline
Activity   & \textbf{563.47} & 1345.61 & \textbf{15729.25}  & 15770.73  & \textbf{2108.1} & 2111.2 \\
\hline
Insulin   & \textbf{181.57} & 190.11 & \textbf{342.74}  &  360.78 & \textbf{55.81} & 61.69 \\
\bottomrule
\end{tabu}}
\end{table}

While ground truth ATEs for the diabetes dataset are unavailable, we validated the detected causal relation shifts against established medical literature. As shown in \cref{tbl:diabetes_ATEestimation}, the shifts identified by our method align with known findings. For example, our model captured the insulin-mediated effect where glucose levels decrease despite increased meal intake, consistent with results reported in \citep{campbell2014metabolic}. Additionally, \cref{tbl:diabetes_processfit} demonstrates that our method outperforms CAUSE in process fitting across all intervention types, underscoring the benefits of incorporating out-of-domain interventions. The model also maintained strong performance in the baseline setting without interventions, confirming its robustness.\looseness-1

\section{Discussion}
While our method offers significant advantages in modeling causal relation shifts under out-of-domain interventions, two key limitations merit consideration. First, our framework simplifies interventions to binary occurrence indicators, potentially limiting expressiveness for complex interventions with varying intensity or duration. Second, if interventions open backdoor causal paths between cause and outcome events, estimation accuracy may be compromised, requiring careful domain knowledge for intervention selection. Scalability concerns are discussed separately in the 
\cref{appendix:scalability}. 
Future work should extend the framework to continuous intervention variables for more fine-grained modeling and improved estimation precision.
\section{Conclusion}

We propose a novel causal framework leveraging out-of-domain interventions to analyze causal relation shifts in temporal process data. Our Transformer-based Hawkes process integrates intervention effects with an unbiased treatment effect estimator for accurate causal quantification. Extensive simulations and real-world validation, corroborated by literature, demonstrated superior performance in both process fitting and ATE estimation, enabling practical extraction of actionable causal knowledge for informed decision-making.
\newpage
\appendix
\section{Proof of Main Theorems}
\label{appendix:proofs}
\subsection{Proof of Theorem 1}
\label{appendix:proofs_thm1}
The definition of direct cause gives that conditional on the direct cause, CIF of $o$ is functionally independent on other variables. 
Recall that we exclude trivial direct cause. 
Thus taking conditional expectation on the direct cause inside the expectation of $\tau (v_t^w)$ completes the proof.

\subsection{Proof of Theorem 2}
\label{appendix:proofs_thm2}
Expanding the estimator $\hat{\tau}(v)$, we get
$
\hat{\tau}(v) = \mathbb{E} [ \frac{1}{T} \int_0^T  (\frac{ \mathbbm{1} \{c_t^w=1, v_t^w=v\}}{e_t^w(v)}  \lambda^{(1,v_t^w)}(t) - \frac{\mathbbm{1} \{c_t^w=0, v_t^w=v\}}{1-e_t^w(v)}\lambda^{(0,v_t^w)}(t)  )dt  ].
$
WLOG, we only justify the component involves $c_t^w=1$. The other component of $c_t^w=0$ follows similarly.   
SUTVA gives 
$$\mathbb{E}   [\int_{0}^T  \frac{ \mathbbm{1} \{c_t^w=1, v_t^w=v\}}{e_t^w(v)} \lambda(t)  dt   ] =  \mathbb{E}  [  \int_{0}^T  \frac{ \mathbbm{1} \{c_t^w=1, v_t^w=v\}}{e_t^w(v)} \lambda^{(1,v)} (t)  dt    ].$$
Taking conditional expectation on $e_t^w(v)$, and combining with  
$$\mathbb{E}  [ \int_{0}^T  \frac{ \mathbbm{1} \{c_t^w=1, v_t^w=v\}}{e_t^w(v)} \lambda^{(1,v)} (t)  dt |e_t^w(v) ]=\mathbb{E}  [ \int_{0}^T   \lambda^{(1,v)} (t)  dt |e_t^w(v) ],$$ that follows overlap and unconfoundedness assumptions, complete the proof.

   \section{Simulating Out-of-Domain interventions}\label{appendix:simulation}

\subsection{Out-of-Domain Intervention Types}
Existing data generation techniques do not support the injection of out-of-domain interventions into temporal process simulation. In our research, we addressed this challenge by modifying the Hawkes process simulation package Tick \citep{bacry2017tick}.

First, a random causal window length \(c_t^w\) is defined for cause event. out-of-domain events \(v\) are generated by randomly selecting cause-outcome pairs \((<c, o>)\) that \(v\) will have its effect upon. Each \(v\) is assigned an occurrence probability \(p_{v}\), forming the out-of-domain intervention vector \(\Omega_{\mathcal{V}} = [{v_1}^{p_{v_1}}, {v_2}^{p_{v_2}},\ldots]\). Additionally, each out-of-domain intervention \(v\) is associated with a randomly generated influence window \(v_t^w\).

During the simulation, the CIF \(\lambda\)  is modified (\(\lambda \rightarrow \lambda'\)) to dynamically inject out-of-domain interventions.  We categorize out-of-domain interventions into three types based on the CIF \ref{eq:CIF} component it effects.

\begin{enumerate}
    \item \textbf{Baseline:}  Modifies the baseline intensity of an outcome event. The baseline intensity is the background rate of the event. (\(\mu_o\rightarrow\mu_{o}'\)).
    \item \textbf{Cause:} Alters the impact of a cause event \(c\) on an outcome event \(o\). (\(\phi_{co}\rightarrow\phi_{co}'\)).
    \item \textbf{Covariate:} Changes the influence of a covariate event \(d\) on an outcome event \(o\). A covariate event modulates the likelihood of the outcome.  (\(\phi_{do}\rightarrow\phi_{do}' : d \neq c\)).
\end{enumerate}

Examples of these out-of-domain intervention types are provided in \ref{subsec:example}. Exact parameters used to generate the experiment datasets are described in \ref{subsec:parameters}. 

Algorithm \ref{alg:ood} describes the process to generate the out-of-domain interventions.
algorithm \ref{alg:dgp} describe the simulation process in detail.

\begin{algorithm}
\caption{Out-of-Domain Intervention Generation}
\label{alg:ood}
\begin{algorithmic}[1]
\Require $c$,$o$,
, $[\phi]$: adjacency, $\vec{\mu}$: baseline 
\Ensure $v_t^{w}$
    \State Randomly Select \((<c, o>)\)
    \State Randomly Select out-of-domain intervention
    \If{type =  \texttt{baseline}}
        \State \(\mu_o \rightarrow \mu_{o}'\)
    \ElsIf{type = \texttt{cause}}
        \State \(\phi_{c,o} \rightarrow \phi'_{c,o}\)
    \ElsIf{type = \texttt{covariate}}
        \State \(\phi_{cov,o} \rightarrow \phi'_{cov,o}\)
    \EndIf
\State \Return $v_t^{w}$
\end{algorithmic}
\end{algorithm}

\begin{algorithm}[h]
\caption{ Out-of-Domain Intervention Injected Simulation}
\label{alg:dgp}
\begin{algorithmic}[1]
\Require 
$v$,
$c$, $o$, $[\phi]$: adjacency, $\vec{\mu}$: baseline 
\Statex $e_t:$ Simulated event at time \(t\)

\While{$t$ \textless $T$}
    \ForAll{$v_t^{w}$}
        \If{\textbf{not} $v$ active}
            \State Randomly choose State
            \If{State = Active}    
                \State set $v$ to active
                \State Add $v_t^{w}$ to  active interventions list, calculate activation end time
                \If{$c$ active} \Comment{Associated cause}
                \State Calculate new CIF $\lambda$ \Comment{Based on out-of-domain intervention type}
                \EndIf
            \EndIf
        \ElsIf {$t \geq $ Activation time end}
                \State set $v$ to inactive
                \State Calculate new CIF $\lambda$ \Comment{Reset}
        \EndIf
    \EndFor
    \State ${e_t} \gets$ \Call{Simulate Hawkes Process}{\null}
    \If{$e_t = c $}
        \State set $c$ to active
        \State  add $c_t^w$ to active cause list and calculate activation end time
    \EndIf  
    \If {$t \geq $ activation time end} 
            \State set $c$ to inactive
        \EndIf
\EndWhile
\State \textbf{Return} 
\end{algorithmic}
\end{algorithm}

\subsection{Out-of-Domain Intervention Examples}
\label{subsec:example}

Consider a scenario with a single cause \(c\)  (\textbf{ticks \#0}), and a corresponding outcome \(o\) \textbf{(ticks \#1)} and a single out-of-domain intervention event \(v\). Suppose, \(c_t^w=0.5\),  \(v_t^w = 0.7\) 

We show the three types of  out-of-domain intervention events based on the scenario above. In each out-of-domain intervention type, cause \(c\) is depicted in the top figure, the outcome \(o\)  is shown in the middle / second figure and the out-of-domain intervention \(v\)  is shown in the bottom figure.

\subsubsection*{Baseline}

The baseline intensity values \(\mu\) for the events are defined as follows:
Cause (Ticks \#0): \(\mu_c = 2.5\)
Outcome (Ticks \#1): \(\mu_o = 1.5\)

The adjacency matrix \(\overrightarrow{\phi}\) is set to zero.

Now, in the given out-of-domain intervention scenario, if the cause window \(c_t^w\), (ticks \#0) and out-of-domain intervention window \(v_t^w\) both are active, the baseline intensity for the outcome event \(o\) (ticks \#1) will increase from \(\mu \leftarrow 1.5\) to \(\mu' \leftarrow 5.5\). This alteration in intensity persists for the duration of the out-of-domain intervention or the cause window, whichever concludes first.

\begin{figure}[] 
\begin{center}
\includegraphics[width=0.5\textwidth]{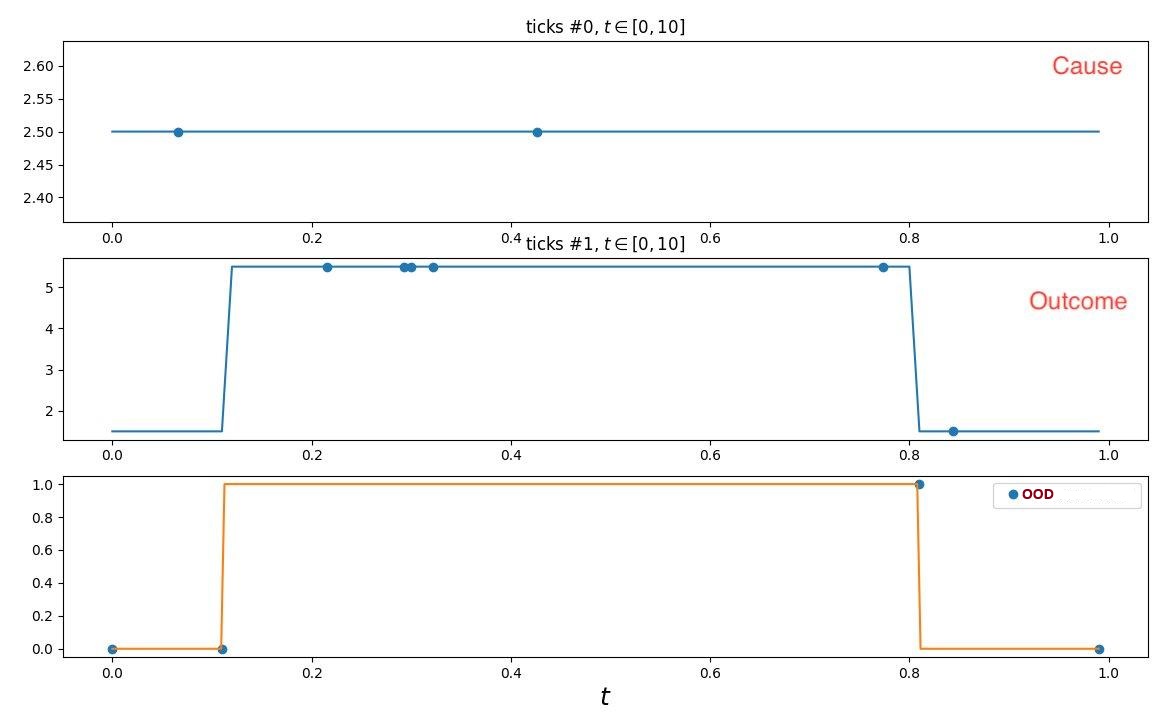}    
\end{center}
\caption{Baseline Out-of-Domain Intervention}
\label{img: baseline}
\end{figure}

The example figure \ref{img: baseline} illustrates the impact of an out-of-domain intervention event on the baseline intensity of the outcome event.

\subsubsection*{Cause}

The baseline intensity values \(\mu\) for the events are defined as follows:
Cause (Ticks \#0): \(\mu_c = 2.5\)
Outcome (Ticks \#1): \(\mu_o = 1.5\)

The adjacency matrix is initialized with zeros \(\overrightarrow{\phi} \leftarrow 0\), signifying that no event is influenced or excited by the appearance of other events within the system.

Now, consider the out-of-domain intervention (bottom figure), the out-of-domain intervention is defined such that, when the out-of-domain intervention  window \(v_t^w\) is active, the adjacency matrix entry for the relationship between outcome and cause will instantaneously jump from \(\phi_{co} \leftarrow 0\) to  \(\phi'_{co} \leftarrow 5\). Moreover, this influence exhibits exponential decay \(e^{-t}\) with a decay factor of \(1\). Consequently, an occurrence of Ticks \#0 will significantly increase the intensity of Ticks \#1 by  5 in presence of out-of-domain intervention. This heightened intensity will persist for the duration of the $v_t^w$ or $c_t^w$, whichever concludes first.

\begin{figure}[h] 
\begin{center}
\includegraphics[width=0.5\textwidth]{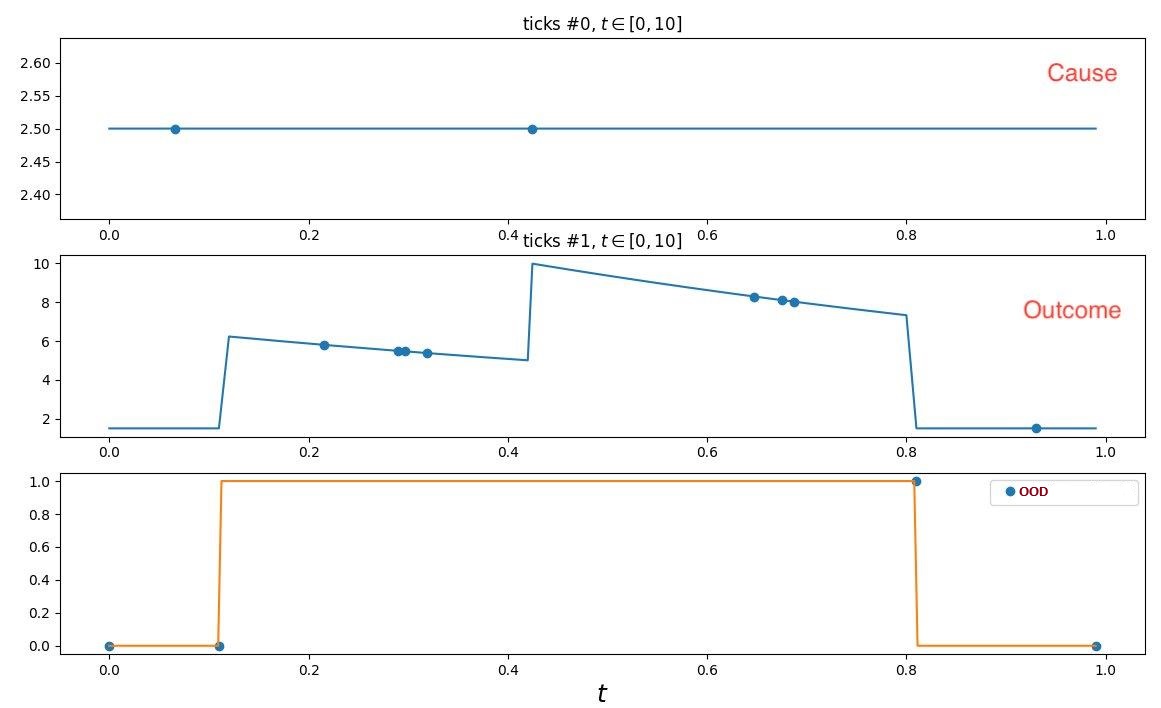}    
\end{center}
\caption{Cause Out-of-Domain Intervention}
\label{img: cause}
\end{figure}

The example figure \ref{img: cause} illustrates this impact.

\subsubsection*{Covariate}

In this scenario, we have an additional covariate event \(d\).
The initial baseline intensity values \(\mu\) for the events are defined as follows:
Cause (Ticks \#0): \(\mu_c = 2.5\)
Outcome (Ticks \#1): \(\mu_o = 1.5\)
Covariate (Ticks \#2): \(\mu_b = 1.5\)

The adjacency matrix \(\overrightarrow{\phi}\) is initialized with zeros, indicating no direct influence between events.

Now, introducing out-of-domain intervention \(v\) (bottom figure), we define it such that when the out-of-domain intervention window \(v_t^w\) and cause window \(c_t^w\) both are active, the adjacency matrix entries for outcome and covariate \(\phi_{do} \leftarrow 0\) will  experience a substantial increase \(\phi_{do'} \leftarrow 7\), characterized by exponential decay \(e^{-t}\) with a factor of \(1\). In practical terms, the occurrence of Ticks \#2 results in a significant intensity boost of Ticks \#1 by \(7\), with this effect sustained for the duration of the $v_t^w$ or $c_t^w$, whichever concludes first.

\begin{figure}[h] 
\begin{center}
\includegraphics[width=0.5\textwidth]{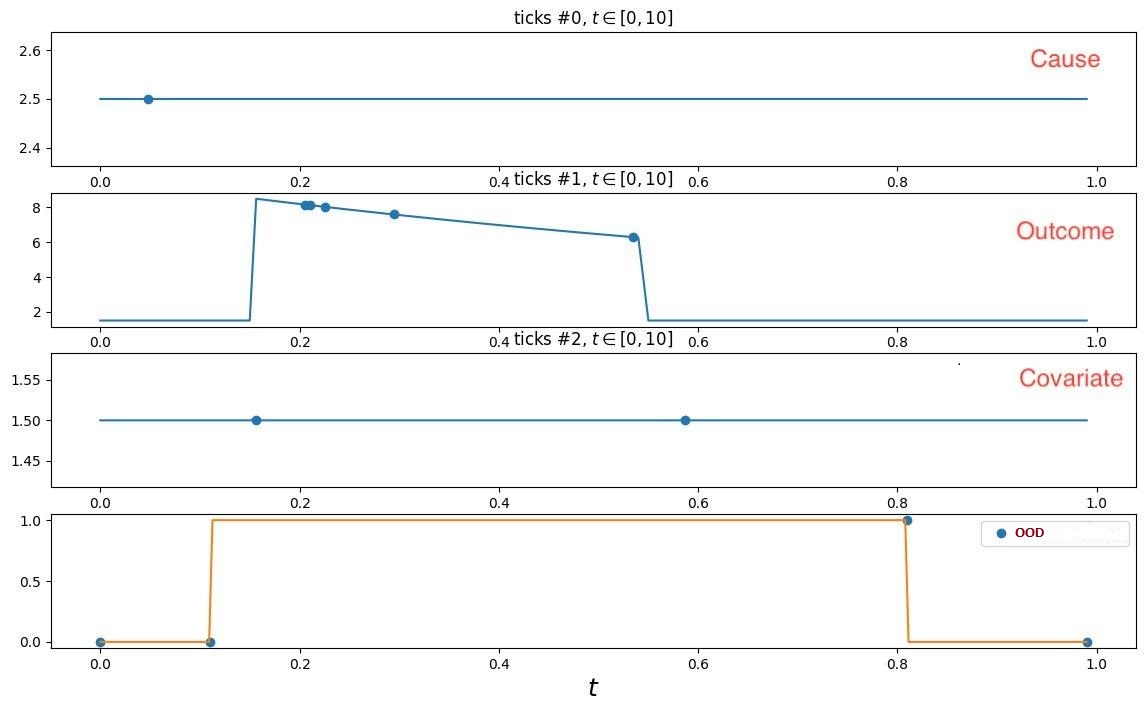}    
\end{center}
\caption{Covariate Out-of-Domain Intervention}
\label{img: covariate}
\end{figure}

This example figure \ref{img: covariate} effectively demonstrates the influence of an out-of-domain intervention on the covariate influence on outcome.

\subsection{Simulation Dataset Parameters}
\label{subsec:parameters}

\begin{table}[h]
    \begin{center}
\caption{Number of Baseline, Cause \& Covariate Out-of-Domain Interventions in Each Simulation Dataset}
\label{tbl:simu_desc}
\arrayrulecolor{lightgray}
    \begin{tabular}{|l|c|c|c|}
    \hline
        \textbf{Dataset} & \textbf{Baseline} &  \textbf{Cause} & \textbf{Covariate}\\
         \hline
         No out-of-domain intervention & 0 & 0 & 0\\
         \hline
         Baseline out-of-domain intervention & 30 & 0 & 0 \\
         \hline
 All out-of-domain intervention & 10 & 12 & 8 \\
 \hline
    \end{tabular}
\end{center}    
\end{table}
To simulate baseline out-of-domain effects, the base intensity rate was increased by a specified percentage (e.g., \(\mu' \sim 150\% \mu\)). 

For cause and covariate out-of-domain intervention impacts, existing causal relationships between events were modified by altering the kernel function \(\phi\). Specifically, for events \(j\) and \(k\) with a causal connection, the corresponding relationship coefficient \((\phi_{jk}=x\)) was set to zero (\(\phi_{jk}'=0\)), indicating a non-causal connection. Conversely, changing the coefficient from zero \((\phi_{jk}=0\)) to a non-zero value \(\phi_{jk}=x\)) represented the introduction of a causal relationship. This dynamic adjustment allowed us to simulate scenarios where causal connections were either disrupted or newly formed due to external out-of-domain intervention influences.

\Cref{tbl:simu_desc} shows the distribution of three out-of-domain intervention types across the simulated datasets.

\section{Model Architecture and Implmentation Details}
\label{appendix:implementation details}


\subsection{Model Architecture Specification}

Our Transformer-based point process model consists of several key components, each designed to capture different aspects of the temporal event sequences with out-of-domain interventions.

\subsubsection*{Embedding Layer:}
The three embeddings below are illustrated in \cref{fig:NNarchitecture}, where event embeddings ($e_i$) and out-of-domain intervention embeddings ($V_i$) are combined via a weighted sum, then merged with the temporal positional encoding ($\Delta t$).

\begin{itemize}
    \item \textbf{Event Embedding}:  
  We convert categorical event types to dense vectors using an embedding lookup table with dimension 128. Each event type (e.g., cause, outcome, other observed events) is assigned a unique learnable embedding vector.
    \item \textbf{Time Embedding}: We encode relative time differences between consecutive events ($t_i - t_{i-1}$) using sinusoidal positional encodings with the same dimension as event embeddings (128). This preserves temporal ordering information while allowing the model to account for varying time intervals.
    
    \item \textbf{out-of-domain Intervention Encoding}:  We represent out-of-domain interventions using a binary vector (not one-hot) where each bit indicates whether a specific intervention type is active (1) or inactive (0) within the time window. The dimension equals the number of distinct intervention types in the dataset.  
    
    Let \(\lVert\mathbb{V}\rVert\) be the number of unique out-of-domain interventions, and \(\mathbb{V}^{bin}\) be the binary vector of length \(||\mathbb{V}||\), \(\mathbb{V}^{bin}_i\) indicating whether the \(i\)th out-of-domain intervention occurred within  \(v_t^w\) in proximal history $H$:
\[
    \mathbb{V}^{bin}_i= 
\begin{cases}
    1,& \text{if } \sum_{t-v_w}^t[v_i=1] \geq 1\\
    0,              & \text{otherwise}
\end{cases}
\]
 
\end{itemize}

\subsubsection*{Transformer Encoder:}
We list the hyperparameters in Transformer Encoder architecture. 

\begin{itemize}
    \item \textbf{Number of Encoder Layers}: 2
    \item \textbf{Number of Attention Heads}: 2
    \item \textbf{Key/Value Dimension per Head}: 64
    \item \textbf{Hidden-State Dimension}: 64 
    \item \textbf{Feed-Forward Inner Dimension}: 256 (4$\times$ model dimension)
    \item \textbf{Dropout Rate}: 0.1
    \item \textbf{Attention Dropout Rate}: 0.1
    \item \textbf{Layer Normalization Epsilon}: $1 \times 10^{-6}$
\end{itemize}

\subsubsection*{Convolutional Neural Network:}
We list the hyperparameters in CNN architecture. 

\begin{itemize}
    \item \textbf{Kernel Size}: 5.  One convolutional layer with multiple output channels, no pooling was used. 
    \item \textbf{Padding}: 2
    \item \textbf{Number of Output Channels}: Equal to number of basis functions
    \item \textbf{Activation Function}: ReLU
\end{itemize}

\subsubsection*{Basis Functions:}
\begin{itemize}
    \item \textbf{Number of Basis Functions (B)}: 8
    \item \textbf{Basis Type}: 1) Unity basis: A constant function that always returns 1.0, 2) Normal basis: 7 Gaussian distributions with dyadic spacing
    \item \textbf{Bandwidth Selection}: Dyadic spacing for maximum mean coverage
\end{itemize}

   \subsection{\textbf{Data Preprocessing}}
    \begin{itemize}
        \item \textbf{Temporal binning:} In handling real-world datasets such as the diabetes dataset with \(||\mathbb{S}||=70\) individuals and the semi-simulated predictive maintenance dataset with \(||\mathbb{S}||=100\) machines, we address the challenge of limited data by employing a technique called temporal binning. For the predictive maintenance dataset, we split the event sequences into subsequences of length \(400\), starting at each \(100^{\text{th}}\) event. Similarly, for the diabetes dataset, we create subsequences of length \(400\) starting at every \(50^{\text{th}}\) event. This approach significantly augments the amount of training samples available for the model.
        \item Normalization of timestamps to [0,1] range
        \item One-hot encoding of event types
        \item Binary encoding of out-of-domain interventions
    \end{itemize}
\subsection{ Hyperparameter Configuration and Tuning}

We implemented our model in Python using PyTorch.  We conducted  hyperparameter search using 5-fold cross-validation on the validation set. \Cref{tab:hyperparameter_config_part1} summarizes the final hyperparameter configuration and the search ranges we explored.

\begin{table}[htbp]
\centering
\caption{Hyperparameter Configuration}
\label{tab:hyperparameter_config_part1}
\begin{tabular}{@{}llll@{}}
\toprule
Hyperparameter & Final Value & Search Range & Selection Method \\
\midrule
Learning Rate & 0.001 & [0.0001, 0.0005, 0.001, 0.005] & Grid Search \\
Batch Size & 64 & [16, 32, 64, 128] & Grid Search \\
Number of Epochs & 300 & Up to 2000 with early stopping & Validation NLL \\
Loss Weight $\alpha$ (CE) & 5.0 & [0.1, 1.0, 5.0, 10.0] & Grid Search \\
Loss Weight $\beta$ (Reg) & 0.01 & [0.001, 0.01, 0.1] & Grid Search \\
Number of Encoder Layers & 2 & [1, 2, 4, 8] & Grid Search \\
Number of Attention Heads & 2 & [1, 2, 4, 8] & Grid Search \\
Embedding Dimension & 128 & [64, 128, 256] & Grid Search \\
Dropout Rate & 0.1 & [0.0, 0.1, 0.01, 0.001] & Grid Search \\
Causal Window Size & 10 & [5, 10, 15, 20] & Grid Search \\
\bottomrule
\end{tabular}
\end{table}

Hyperparameter selection was based on validation set negative log-likelihout-of-domain (NLL), with early stopping patience of $20$ epochs. We found that increasing model complexity beyond our final configuration (e.g., more layers or attention heads) did not significantly improve performance while substantially increasing computational cost.

\subsection{Training Procedure}

Our training methodology follows these steps:

\begin{enumerate}
 
    \item \textbf{Optimization Strategy}:
    \begin{itemize}
        \item Optimizer: Adam ($\beta_1=0.9, \beta_2=0.999, \epsilon=1 \times 10^{-8}$)
        \item Learning Rate Schedule: Triangular cyclic learning rate with \texttt{base\_lr=0.0005}, \texttt{max\_lr=0.001}, \texttt{step\_size=20} epochs
        \item Gradient Clipping: Maximum norm of 1.0 to prevent exploding gradients
        \item Early Stopping: Monitor validation NLL with patience=20 epochs
    \end{itemize}
    \item \textbf{Training Dynamics}:
    \begin{itemize}
        \item Loss components are weighted as per Equation 6: $L = L_{\text{NLL}} + \alpha L_{\text{CE}} + \beta L_{\text{reg}}$
        \item Cross-entropy weight $\alpha=5.0$ emphasizes accurate event type prediction
        \item Regularization weight $\beta=0.01$ prevents overfitting while allowing model flexibility
    \end{itemize}
    \item \textbf{Batch Processing}:
    \begin{itemize}
        \item Sequences padded to the maximum length within each batch
        \item Masked attention used to ignore padding tokens
    \end{itemize}
\end{enumerate}

\subsection{Computational Requirements}

Our implementation was developed using PyTorch 2.10.0. Training and evaluation were conducted with the following computational resources:

\begin{itemize}
    \item \textbf{Hardware}: NVIDIA V100 GPU with 16GB memory
    \item \textbf{Training Time}:
    \begin{itemize}
        \item Simulated Data (1000 sequences): $\sim$3 hours for full training
        \item Diabetes Dataset (70 patients): $\sim$1 hour for full training
        \item Predictive Maintenance (100 machines): $\sim$8 hours for full training
    \end{itemize}
\end{itemize}

\section{Comparison with State-of-the-Art Methods}
\label{appendix:SOTA}

Our framework addresses limitations in existing approaches by explicitly modeling how out-of-domain interventions affect causal relationships in temporal point processes. Below, we compare our approach with state-of-the-art methods across relevant categories and analyze their limitations in handling out-of-domain interventions.

\subsection{Causal Discovery Methods for Event Sequences}

The Temporal Causal Discovery Framework (TCDF) \citep{nauta2019causal} employs attention-based convolutional neural networks to identify time-delayed causal relationships in time series data without requiring prior knowledge. While TCDF provides interpretable attention scores for causal discovery, it is primarily designed for regularly-sampled multivariate time series rather than point processes with irregular timestamps.

CAUSE (CAusal Understanding in Search Explanations) \citep{zhang2020cause} learns Granger causality from event sequences using attribution methods such as integrated gradients. Although CAUSE is  designed for event sequences and effectively accounts for event history when quantifying causal influence, it focuses on Granger causality—which measures prediction improvement rather than true pairwise causality. 

Similarly, the Self-Explaining Neural Networks approach for Granger causality \citep{marcinkevivcs2021interpretable} provides interpretable feature attributions and handles nonlinear relationships in temporal data. However, being based on the Granger causality framework, it doesn't capture pairwise intervention-based causality. 

\subsection{Neural Temporal Point Process Models}

The Neural Hawkes Process \citep{mei2017neural} uses continuous-time LSTM to modulate intensity functions based on event history, capturing complex temporal dependencies and allowing for flexible intensity functions. Despite these strengths, this approach includes no causal inference framework.

Building on this, the Transformer Hawkes Process \citep{zuo2020transformer} applies transformer architecture with self-attention to model event dependencies, effectively capturing long-range dependencies and handling variable-length sequences efficiently. However, like its predecessors, it provides no mechanism to identify causal inferences.

The Temporal Attention Augmented Transformer Hawkes Process \citep{zhang2022temporal} enhances the transformer architecture with temporal attention specifically for point processes, incorporating temporal information explicitly and improving prediction accuracy. While innovative in its modeling approach, this method focuses on prediction rather than causal inference,
and cannot quantify how causal relationships change under different intervention scenarios.

\subsection{Treatment Effect Estimation Approaches}

Counterfactual Temporal Point Processes \citep{noorbakhsh2022counterfactual} generates counterfactual realizations of temporal point processes using marked point processes, providing a framework for counterfactual reasoning in point processes and handling complex event dependencies. This approach, however, focuses on counterfactual predictions rather than identifying causal relation shifts.

The Causal Inference framework for Event Pairs in Multivariate Point Processes \citep{gao2021causal} uses inverse probability weighting to estimate causal effects between event pairs, addressing confounding in point processes through propensity score adjustment. While methodologically related to our approach, this framework does not consider external interventions that may modify causal relationships. It assumes static causal relationships across the entire process and cannot identify how treatment effects vary under different intervention scenarios.

The Causal Transformer for Estimating Counterfactual Outcomes \citep{melnychuk2022causal} uses transformer architecture to estimate counterfactual outcomes from patient trajectories, effectively handling temporal dependencies in treatment effect estimation and accounting for time-varying confounders. However, this method is designed for regularly-sampled time series rather than point processes with irregular timestamps. 

\subsection{Advantages of Our Approach}

Our proposed method addresses these limitations by:
\begin{enumerate}
    \item Explicitly incorporating out-of-domain interventions into the temporal point process framework with a specialized embedding mechanism
    \item Developing a theoretical foundation to quantify causal relation shifts under different intervention states
    \item Introducing a novel neural architecture that simultaneously captures event dependencies and intervention effects
    \item Providing an unbiased estimation method for intervention-conditional treatment effects using adaptive propensity score adjustment
\end{enumerate}

This integrated approach enables the identification of how causal relationships between events change under different out-of-domain intervention scenarios—a capability not present in existing methods.
\section{Scalability}
\label{appendix:scalability}
\subsection{Scalability Limitation}

Several challenges may arise when scaling our OOD causal framework to larger datasets and more complex OOD scenarios. 

\begin{enumerate}    
    \item \textbf{Intervention Combinatorial Explosion:} As the number of potential OOD intervention types increases, the number of possible intervention combinations grows exponentially, creating data sparsity issues when estimating propensity scores for rare intervention patterns.
    
    \item \textbf{Memory Requirements for Basis Functions:} Our basis function approach requires maintaining separate weights for each event type and basis function combination, which scales linearly with the product of these dimensions. For domains with hundreds of event types, memory constraints become significant.
    
    \item \textbf{Propensity Score Estimation:} The accuracy of propensity score estimation deteriorates with increasing dimensionality of the covariate space, which is exacerbated when many event types and intervention combinations must be considered simultaneously.

    \item \textbf{Transformer Complexity:} The self-attention mechanism in our Transformer encoder has quadratic complexity with respect to sequence length. It creates computational bottlenecks for very long event sequences. 
\end{enumerate}

\vspace{-4mm}
\subsection{Future Directions to Mitigate Scalability Concerns}
\vspace{-2mm}

We propose several promising directions could address these OOD causal framwork and scalability concerns. 
\begin{enumerate}[noitemsep, topsep=1pt]
    \item \textbf{Hierarchical Intervention Modeling:} Grouping related interventions and modeling their effects hierarchically could mitigate the combinatorial explosion problem while preserving the ability to detect important causal shifts.
    \item \textbf{Efficient Transformer Variants:} Adaptations like KV Cache \cite{kwon2023efficient} and FlashAttention \cite{dao2022flashattention} could reduce the computational burden of the Transformer component while maintaining modeling capacity for intervention effects.
\end{enumerate}

\bibliography{references}
\bibliographystyle{splncs04}

\end{document}